\documentclass[runningheads]{llncs}
\usepackage[T1]{fontenc}
\usepackage{bm} 
\usepackage{graphicx}
\usepackage{mathrsfs}
\usepackage{array}
\usepackage{amsfonts, amssymb}
\usepackage{amsmath}
\usepackage[mathscr]{eucal}
\usepackage{mathtools}
\usepackage{diagbox}
\usepackage{float}
\usepackage{bbding}
\usepackage{xcolor}
\usepackage{hyperref}
\usepackage{ulem}
\usepackage{makecell}
\usepackage{dsfont}
\usepackage{multirow}
\usepackage{cite}
\usepackage{ulem}
\usepackage{booktabs}
\usepackage{wrapfig}

\newcommand{\bfx}{\mathbf{x}}

\newcommand{\bfc}{\mathbf{c}}

\newcommand{\bbE}{\mathbb{E}}
\newcommand{\calN}{\mathcal{N}}

\newcommand{\calD}{\mathcal{D}}
\newcommand{\calL}{\mathcal{L}}

\newcommand{\calR}{\mathcal{R}}

\newcommand{\calM}{\mathcal{M}}

\newcommand{\calB}{\mathcal{B}}
\newcommand{\calU}{\mathcal{U}}
\newcommand{\calS}{\mathcal{S}}
\newcommand{\bfe}{\mathbf{e}}

\newcommand{\bfp}{\mathbf{p}}

\usepackage{CJKutf8}
\usepackage{changes}

\usepackage{algorithm}
\usepackage{algpseudocode}
\usepackage{stfloats}
\usepackage{tabularx}
\usepackage{colortbl}
\usepackage{pifont}

\makeatletter

\newcommand{\Rmnum}[1]{\expandafter\@slowromancap\romannumeral #1@}
\makeatother

\begin{document}
\title{MedQ-Engine: A Closed-Loop Data Engine for Evolving MLLMs in Medical Image Quality Assessment}

\titlerunning{MedQ-Engine: Data Engine for Medical IQA}

\author{Jiyao Liu\inst{1}$^\dagger$ \and
Junzhi Ning \inst{2}$^\dagger$  \and 
Wanying Qu \inst{1}  \and 
Lihao Liu\inst{2}  \and 
Chenglong Ma\inst{1}  \and  \\
Junjun He \inst{2}$^{\text{\Envelope}}$  \and 
Ningsheng Xu\inst{1}$^{\text{\Envelope}}$
}

\authorrunning{J Liu et al. }

\institute{Fudan University, Shanghai, China \and
Shanghai Artificial Intelligence Laboratory, Shanghai, China}

\maketitle              

\renewcommand\thefootnote{}
\footnotetext{$^\dagger$Equal contribution. $^{\text{\Envelope}}$Corresponding author. }
\renewcommand\thefootnote{\arabic{footnote}}

\begin{abstract}
Medical image quality assessment (Med-IQA) is a prerequisite for clinical AI deployment, yet multimodal large language models (MLLMs) still fall substantially short of human experts, particularly when required to provide descriptive assessments with clinical reasoning beyond simple quality scores. However, improving them is hindered by the high cost of acquiring descriptive annotations, and by the inability of one-time data collection to adapt to the model's evolving weaknesses. To address these challenges, we propose \textbf{MedQ-Engine}, a closed-loop data engine that iteratively \textit{evaluates} the model to discover failure prototypes via data-driven clustering, \textit{explores} a million-scale image pool using these prototypes as retrieval anchors with progressive human-in-the-loop annotation, and \textit{evolves} through quality-assured fine-tuning, forming a self-improving cycle. Models are evaluated on complementary \textit{perception} and \textit{description} tasks. An entropy-guided routing mechanism triages annotations to minimize labeling cost. Experiments across five medical imaging modalities show that MedQ-Engine elevates an 8B-parameter model to surpass GPT-4o by over 13\% and narrow the gap with human experts to only 4.34\%, using only 10K annotations with more than 4$\times$ sample efficiency over random sampling.

\end{abstract}

\section{Introduction}
\label{sec:intro}

\begin{figure}[t!]
  \centering
  \includegraphics[width=\textwidth]{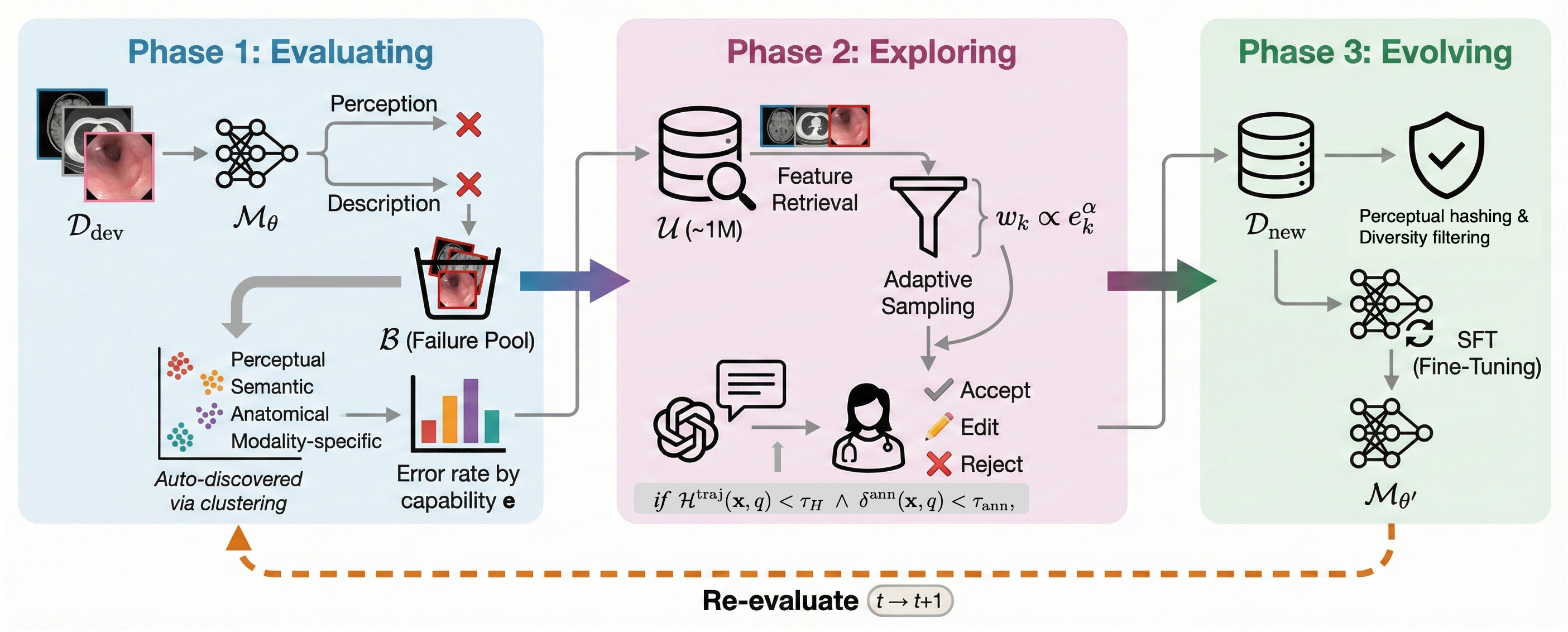}
  \caption{\textbf{Overview of MedQ-Engine.} Our closed-loop data engine iteratively improves MLLMs for Med-IQA via three phases: \textit{Evaluating} (failure clustering on a dev set), \textit{Exploring} (prototype-based retrieval and human-in-the-loop annotation), and \textit{Evolving} (quality-assured fine-tuning with re-evaluation).}
  \label{fig:framework}
  \vspace{-1.5em}
\end{figure}

Medical image quality assessment (Med-IQA) is essential for reliable diagnosis, yet the heterogeneity of modalities, anatomical regions, and clinical scenarios poses significant challenges~\cite{xun2025mediqa}. Existing approaches either produce modality-agnostic scalar scores~\cite{mittal2012no,xun2025mediqa} or are confined to specific modalities~\cite{chen2024iqagpt,xun2024chest}. Multimodal large language models (MLLMs) offer a promising cross-modality alternative, capable of generating descriptive assessments that identify degradation types, analyze visual impact, and evaluate degradation severity~\cite{zhang2024q,you2024depicting}. Yet recent benchmarking reveals that MLLMs still exhibit substantial deficiencies in Med-IQA, with systematic performance gaps compared to human experts~\cite{liu2025medq}. Notably, MedQ-Bench~\cite{liu2025medq} shows that \textit{MLLM errors concentrate in specific capability-modality intersections rather than being uniformly distributed}, suggesting that targeted remediation is far more efficient than uniform data augmentation. However, translating this insight into practice faces two obstacles: a \textit{cost-value dilemma} where simple scoring provides minimal training signal while comprehensive expert descriptions are prohibitively expensive, and \textit{static data collection} that cannot adapt to evolving model weaknesses as new bottlenecks emerge after each improvement.

To address these challenges, we propose \textbf{MedQ-Engine}, a closed-loop data engine that systematically improves MLLM capabilities for Med-IQA through three iterative phases (Fig.~\ref{fig:framework}): (1) \textit{Evaluating}, which assesses the model on a dedicated development set and clusters failure cases into \textit{failure prototypes} that capture dominant error patterns; (2) \textit{Exploring}, which uses these prototypes as retrieval anchors to expand training data from a large-scale image pool ($\sim$1M images), combined with human-in-the-loop verification of AI-generated pre-annotations; (3) \textit{Evolving}, which fine-tunes on quality-assured data and re-evaluates, forming a closed loop that progressively resolves model weaknesses.

\textbf{Contributions.} (1) We propose MedQ-Engine, the first closed-loop data engine for Med-IQA that transforms data-driven error analysis into systematic model improvement through iterative evaluate-explore-evolve phases. (2) We introduce a data-driven failure discovery mechanism with error-weighted adaptive sampling, combined with a human-in-the-loop annotation paradigm that maximizes information gain per expert minute. (3) Extensive experiments and ablation studies across five medical imaging modalities demonstrate that MedQ-Engine substantially narrows the gap with human experts using only 10K samples.

\section{Method}
\label{sec:method}

\subsection{Problem Formulation}
\label{sec:formulation}

Following~\cite{liu2025medq}, we formulate Med-IQA as two complementary tasks. \textit{Perception task} evaluates quality-related visual perception through multi-choice questions, including: (1) \textit{Yes-or-No} questions for binary quality classification (e.g., ``Does this image contain artifacts?''), (2) \textit{What} questions for multi-class identification of degradation types, and (3) \textit{How} questions for severity assessment. \textit{Description task} requires generating comprehensive responses that cover modality and anatomy identification, degradation characterization, technical attribution, visual impact assessment, and an overall quality judgment.

Let $\calD_\text{test}$ denote the evaluation benchmark and $\calU$ denote the unlabeled image pool ($|\calU| \approx 10^6$). We additionally construct a development set $\calD_\text{dev}$, held out independently from $\calD_\text{test}$, to enable iterative failure analysis without data leakage. Our goal is to select a training subset $\calD_\text{train} \subset \calU$ under annotation budget $B$ that maximally improves $\calM_\theta$ (an autoregressive LLM policy parameterized by $\theta$) on $\calD_\text{test}$. Since exhaustive search over this combinatorial space is intractable, MedQ-Engine approximates the solution through an iterative evaluate-explore-evolve loop that progressively identifies and resolves the most impactful failure modes.

\subsection{Phase 1: Evaluating}
\label{sec:evaluating}

\textbf{Failure Case Collection.} We evaluate $\calM_\theta$ on $\calD_\text{dev}$ $R$ times across perception and description tasks spanning multiple modalities. A sample is identified as a failure case and added to the failure pool $\calB$ when its error rate across $R$ runs exceeds a threshold $\gamma$, indicating a persistent weakness rather than stochastic variance. Each failure case is associated with a capability dimension label vector $\bfc = (c_1, \ldots, c_K)$.

\textbf{Data-driven Failure Clustering.} Rather than imposing predefined error categories, we characterize failure patterns directly from model behavior. For each failure case in $\calB$, we represent it as a feature vector that combines its visual content and question-answer information. We then apply agglomerative clustering on these feature vectors and select the number of clusters via the silhouette criterion, yielding $N_c$ cluster centroids $\{\bfp_1, \ldots, \bfp_{N_c}\}$ that serve as \textit{failure prototypes}. In Phase~2, only the visual component of each prototype is used as a retrieval anchor to query the unlabeled image pool.

\textbf{Capability Dimension Analysis.} We aggregate $\calB$ by capability dimensions to compute error rate distribution:
\begin{equation}
    \bfe = (e_1, \ldots, e_K), \quad e_k = \frac{|\{b \in \calB : c_k(b) = 1\}|}{|\{s \in \calD_\text{dev} : c_k(s) = 1\}|},
    \label{eq:error_dist}
\end{equation}
which reveals systematic model weaknesses and guides subsequent data collection.

\subsection{Phase 2: Exploring}
\label{sec:exploring}

\textbf{Prototype-based Retrieval.} We construct the $\sim$1M images pool $\calU$ with  five modalities, sourced from FastMRI~\cite{zbontar2018fastmri}, AAPM CT-MAR~\cite{haneda2025aapm}, MR-ART~\cite{narai2022movement}, HistoArtifacts~\cite{fuchs2024harp}, EndoCV2020~\cite{polat2020endoscopic}, EyeQ~\cite{fu2019evaluation}, GMAI-MMBench~\cite{ye2024gmai}, OmniMedVQA~\cite{hu2024omnimedvqa}, and Radiopaedia\footnote{\url{https://radiopaedia.org/}}, and encoded using BiomedCLIP~\cite{zhang2025multimodal}. Rather than retrieving per individual failure case, we use the $N_c$ failure prototypes from Phase~1 as query anchors. Since each prototype is derived from image and Q-A features jointly, we extract its visual component $\bfp_j^{\text{vis}}$ for retrieval against the image-only embeddings in $\calU$:
\begin{equation}
    \calN(\bfp_j) = \{\bfx \in \calU : \cos(\bfp_j^{\text{vis}}, f_\text{enc}(\bfx)) > \tau_\text{sim}\}.
    \label{eq:retrieval}
\end{equation}
Adaptive sampling weights $w_k \propto e_k^\alpha$ prioritize weak capability dimensions, forming annotation set $\calS$ with $|\calS| = B$.

\textbf{Progressive Human-in-the-loop Annotation.} We design a progressive strategy that adapts human effort to annotation difficulty across iterations, substantially reducing expert cost as the model evolves.

\textit{Cold start ($t{=}0$):} GPT-4o~\cite{achiam2023gpt} pre-annotates the initial 2K samples and domain experts review all annotations via Accept / Edit / Reject, building a high-quality seed dataset $\calD_{\text{new}}^{(0)}$.

\textit{Self-evolution ($t{>}0$):} For each new sample, we obtain both a self-annotation $\hat{y}^{\text{self}}$ from $\calM_{\theta^{(t)}}$ and a reference annotation $\hat{y}^{\text{GPT}}$ from GPT-4o, then route it into one of three paths based on two signals: (i) model confidence, measured by trajectory-level entropy~\cite{agarwal2024unreasonable}:
\begin{equation}
    \mathcal{H}^{\text{traj}}(\bfx,q) = -\frac{1}{|\hat{y}^{\text{self}}|} \sum_{l=1}^{|\hat{y}^{\text{self}}|} \log \calM_{\theta^{(t)}}(\hat{y}^{\text{self}}_l \mid \bfx, q, \hat{y}^{\text{self}}_{<l}),
    \label{eq:entropy}
\end{equation}
and (ii) oracle-agreement $\delta^{\text{ann}} \triangleq \calR(\hat{y}^{\text{self}}, \hat{y}^{\text{GPT}})$. The routing logic is: \textbf{(a)} if the model is \textit{uncertain} ($\mathcal{H}^{\text{traj}} \geq \tau_H$), adopt $\hat{y}^{\text{GPT}}$; \textbf{(b)} if \textit{confident yet disagrees} with the oracle ($\mathcal{H}^{\text{traj}} < \tau_H \wedge \delta^{\text{ann}} < \tau_{\text{ann}}$), escalate to expert review; \textbf{(c)} if \textit{confident and consistent}, adopt $\hat{y}^{\text{self}}$ directly. 

\subsection{Phase 3: Evolving}
\label{sec:evolving}

\textbf{Quality Assurance.} We ensure clinical reliability through: (1) \textit{Deduplication} via perceptual hashing~\cite{zauner2010implementation}; (2) \textit{Diversity filtering} using TF-IDF~\cite{ramos2003using} thresholds to remove near-duplicate descriptions.

\textbf{Model Fine-tuning.} We fine-tune $\calM_\theta$ using supervised instruction tuning with full parameter updating:
\begin{equation}
    \calL_\text{SFT} = -\bbE_{(\bfx,q,y)\sim\calD_\text{new}} \left[\sum_{i=1}^{|y|} \log p_\theta(y_i \mid y_{<i}, \bfx, q)\right].
    \label{eq:sft}
\end{equation}

\textbf{Closed-loop Iteration.} After fine-tuning, the updated model $\calM_{\theta'}$ re-enters the Evaluating phase, updating failure pool $\calB^{(t+1)}$ and error distribution $\bfe^{(t+1)}$. Iteration terminates when performance on $\calD_\text{dev}$ plateaus.

\section{Experiments}
\label{sec:experiment}

\subsection{Experimental Setup}

\textbf{Datasets.} We construct our unlabeled image pool $\calU$ from five medical imaging modalities: MRI, CT, endoscopy, fundus photography, and histopathology, totaling approximately 1 million images from public repositories. We additionally curate a development set $\calD_\text{dev}$ ($\sim$2k samples) for iterative failure analysis, which is held out independently from the test benchmark to prevent data leakage. To ensure strict data integrity, we verify that $\calU$, $\calD_\text{dev}$, and $\calD_\text{test}$ are sourced from disjoint patient cohorts and scanning sessions; perceptual-hash deduplication confirms zero image overlap across all three splits. The Med-IQA abilities of MLLMs after instruction tuning are quantitatively evaluated on MedQ-Bench~\cite{liu2025medq} in two complementary tasks: \textit{Perception}, by measuring the accuracy of answering multi-choice questions (MCQ) related to image quality attributes; and \textit{Description}, which examines how MLLMs can describe image quality issues in natural language.

\textbf{Evaluation Metrics.} For \textit{Perception}, we report type accuracy measuring correct identification of degradation categories. For \textit{Description}, we employ four expert-evaluated dimensions: Completeness, assessing coverage of quality issues; Preciseness, measuring description accuracy; Consistency, evaluating alignment between analysis and conclusion; and Quality Accuracy, verifying correctness of clinical impact assessment. 

\textbf{Comparison Methods.} We compare against state-of-the-art MLLMs spanning four categories: (1) \textit{Open-source general MLLMs}: Qwen2.5-VL-Instruct (7B/32B/72B)~\cite{bai2023qwenvl} and InternVL3 (8B/38B)~\cite{chen2024internvl}; (2) \textit{Medical-specialized MLLMs}: BiMediX2~\cite{mullappilly2024bimedix2}, Lingshu~\cite{xu2025lingshu}, and MedGemma~\cite{sellergren2025medgemma}; (3) \textit{Closed-source MLLMs}: Mistral-Medium-3~\cite{jiang2023mistral7b}, Claude-4-Sonnet~\cite{anthropic2025claude4systemcard}, Gemini-2.5-Pro~\cite{comanici2025gemini}, Grok-4~\cite{xai2025grok4}, and GPT-4o~\cite{achiam2023gpt}. We also include human expert and non-expert performance as reference baselines.

\textbf{Implementation Details.} We use Qwen2.5-VL-7B~\cite{bai2023qwenvl} and InternVL3-8B~\cite{chen2024internvl} as base models $\calM_\theta$, with GPT-4o~\cite{achiam2023gpt} as the annotation oracle. We set similarity threshold $\tau_\text{sim}{=}0.75$, evaluation runs $R{=}5$, and failure frequency threshold $\gamma{=}0.6$. Fine-tuning uses learning rate $2\times10^{-5}$ with cosine schedule over 3 epochs. The annotation budget per iteration is $B{=}2000$ samples, with the first iteration ($t{=}0$) applying full human review and subsequent iterations applying entropy-guided selective review. All experiments are conducted on 8$\times$A100 GPUs.

\subsection{Main Results}

\begin{table*}[t]
    \centering
    \caption{Performance comparison on the MedQbench+ benchmark. \colorbox{green!8}{Green}: open-source general MLLMs; \colorbox{blue!8}{Blue}: medical-specialized MLLMs; \colorbox{orange!15}{Orange}: closed-source MLLMs; \colorbox{yellow!20}{Yellow}: after MedQ-Engine optimization.}
    \label{tab:main_results}
    \setlength{\tabcolsep}{4pt}
    \resizebox{\textwidth}{!}{
    \begin{tabular}{lccccccccc}
    \toprule
    \textbf{Sub-categories} & \multicolumn{4}{c}{\textbf{Perception}} & \multicolumn{5}{c}{\textbf{Description}} \\
    \cmidrule(lr){2-5} \cmidrule(lr){6-10}
    \textbf{Model} & \textbf{Yes/No$\uparrow$} & \textbf{What$\uparrow$} & \textbf{How$\uparrow$} & \textbf{Overall$\uparrow$} & \textbf{Comp.$\uparrow$} & \textbf{Prec.$\uparrow$} & \textbf{Cons.$\uparrow$} & \textbf{Qual.$\uparrow$} & \textbf{Overall$\uparrow$} \\
    \midrule
    \textit{random guess} & 50.00 & 28.48 & 33.30 & 37.94 &  &  &  &  &  \\
    \rowcolor{gray!15} \textit{Non-experts} & 67.50 & 57.50 & 57.50 & 62.50 & - & - & - & - & - \\
    \rowcolor{gray!15} \textit{Human experts} & 88.50 & 77.50 & 77.50 & 82.50 & - & - & - & - & - \\
    \midrule
    \rowcolor{blue!8} BiMediX2-8B~\cite{mullappilly2024bimedix2}        & 44.98 & 27.52 & 27.81 & 35.10 & 0.376 & 0.394 & 0.281 & 0.670 & 1.721 \\
    \rowcolor{blue!8} Lingshu-32B~\cite{xu2025lingshu}        & 50.36 & 50.39 & 51.74 & 50.88 & 0.624 & 0.697 & 1.932 & 1.059 & 4.312 \\
    \rowcolor{blue!8} MedGemma-27B~\cite{sellergren2025medgemma}       & 67.03 & 48.06 & 50.72 & 57.16 & 0.742 & 0.471 & 1.579 & 1.262 & 4.054 \\
    \rowcolor{orange!15} Mistral-Medium-3~\cite{jiang2023mistral7b} & 65.95 & 48.84 & 52.97 & 57.70 & 0.923 & 0.729 & 1.566 & 1.339 & 4.557 \\
    \rowcolor{green!8} Qwen2.5-VL-32B~\cite{wang2024qwen2}    & 67.38 & 43.02 & 58.69 & 59.31 & 1.077 & 0.928 & \textbf{1.977} & 1.290 & 5.272 \\
    \rowcolor{orange!15} Claude-4-Sonnet~\cite{anthropic2025claude4systemcard}  & 71.51 & 46.51 & 54.60 & 60.23 & 0.742 & 0.633 & 1.778 & 1.376 & 4.529 \\
    \rowcolor{green!8} InternVL3-38B~\cite{chen2024internvl}     & 69.71 & 57.36 & 52.97 & 61.00 & 0.964 & 0.824 & 1.860 & 1.317 & 4.965 \\
    \rowcolor{orange!15} Gemini-2.5-Pro~\cite{comanici2025gemini}   & 75.13 & 55.02 & 50.54 & 61.88 & 0.878 & 0.891 & 1.688 & 1.561 & 5.018 \\
    \rowcolor{green!8} Qwen2.5-VL-72B~\cite{wang2024qwen2}    & 78.67 & 42.25 & 56.44 & 63.14 & 0.905 & 0.860 & 1.896 & 1.321 & 4.982 \\
    \rowcolor{orange!15} Grok-4~\cite{xai2025grok4}           & 73.30 & 48.84 & 59.10 & 63.14 & 0.982 & 0.846 & 1.801 & 1.389 & 5.017 \\
    \rowcolor{orange!15} GPT-4o~\cite{achiam2023gpt}           & 78.48 & 49.64 & 57.32 & 64.79 & 1.009 & 1.027 & 1.878 & 1.407 & 5.321 \\
    \midrule
    \rowcolor{green!8} Qwen2.5-VL-7B~\cite{wang2024qwen2}     & 57.89 & 48.45 & 54.40 & 54.71 & 0.715 & 0.670 & \textbf{1.855} & 1.127 & 4.367 \\
    \rowcolor{yellow!20} \makecell[l]{Qwen2.5-VL-7B-10k$^\dagger$\\(Ours)} & \makecell{86.56\\{\scriptsize\color{red}+28.67}} & \makecell{68.22\\{\scriptsize\color{red}+19.77}} & \makecell{62.78\\{\scriptsize\color{red}+8.38}} & \makecell{74.02\\{\scriptsize\color{red}+19.31}} & \makecell{0.950\\{\scriptsize\color{red}+0.24}} & \makecell{0.937\\{\scriptsize\color{red}+0.27}} & \makecell{1.839\\{\scriptsize\color{blue}-0.016}} & \makecell{1.520\\{\scriptsize\color{red}+0.39}} & \makecell{5.246\\{\scriptsize\color{red}+0.88}} \\
    \rowcolor{green!8} InternVL3-8B~\cite{chen2024internvl}      & 72.04 & 47.67 & 52.97 & 60.08 & 0.928 & 0.878 & \textbf{1.858} & 1.317 & 4.983 \\
    \rowcolor{yellow!20} \makecell[l]{InternVL3-8B-10k$^\dagger$\\(Ours)} & \makecell{\textbf{89.61}\\{\scriptsize\color{red}+17.57}} & \makecell{\textbf{73.64}\\{\scriptsize\color{red}+25.97}} & \makecell{\textbf{67.48}\\{\scriptsize\color{red}+14.51}} & \makecell{\textbf{78.16}\\{\scriptsize\color{red}+18.08}} & \makecell{\textbf{1.054}\\{\scriptsize\color{red}+0.13}} & \makecell{\textbf{1.091}\\{\scriptsize\color{red}+0.21}} & \makecell{1.846\\{\scriptsize\color{blue}-0.012}} & \makecell{\textbf{1.661}\\{\scriptsize\color{red}+0.34}} & \makecell{\textbf{5.652}\\{\scriptsize\color{red}+0.67}} \\
    \bottomrule
    \end{tabular}}
    \vspace{-5pt}
\end{table*}

Table~\ref{tab:main_results} presents benchmark results on MedQbench, from which several findings emerge. \textbf{First}, our optimized InternVL3-8B-10k achieves 78.16\% overall perception accuracy, surpassing GPT-4o (64.79\%) by over 13 percentage points and narrowing the gap with human experts (82.50\%) to only 4.34\%. \textbf{Second}, both 7B/8B-scale optimized models consistently outperform all larger counterparts, including 32B, 72B, and closed-source models, demonstrating that targeted data curation can compensate for orders-of-magnitude differences in model scale. \textbf{Third}, medical-specialized MLLMs underperform general-purpose models of comparable size, suggesting that existing medical pretraining does not transfer effectively to the quality assessment domain. On the description task, both optimized models achieve the highest overall scores, with notable improvements in Completeness and Quality Accuracy.

\subsection{Analysis}

We examine MedQ-Engine along four dimensions: qualitative improvements, sample efficiency, component contributions, and annotation cost.

\begin{figure*}[t]
    \centering
    \includegraphics[width=\textwidth]{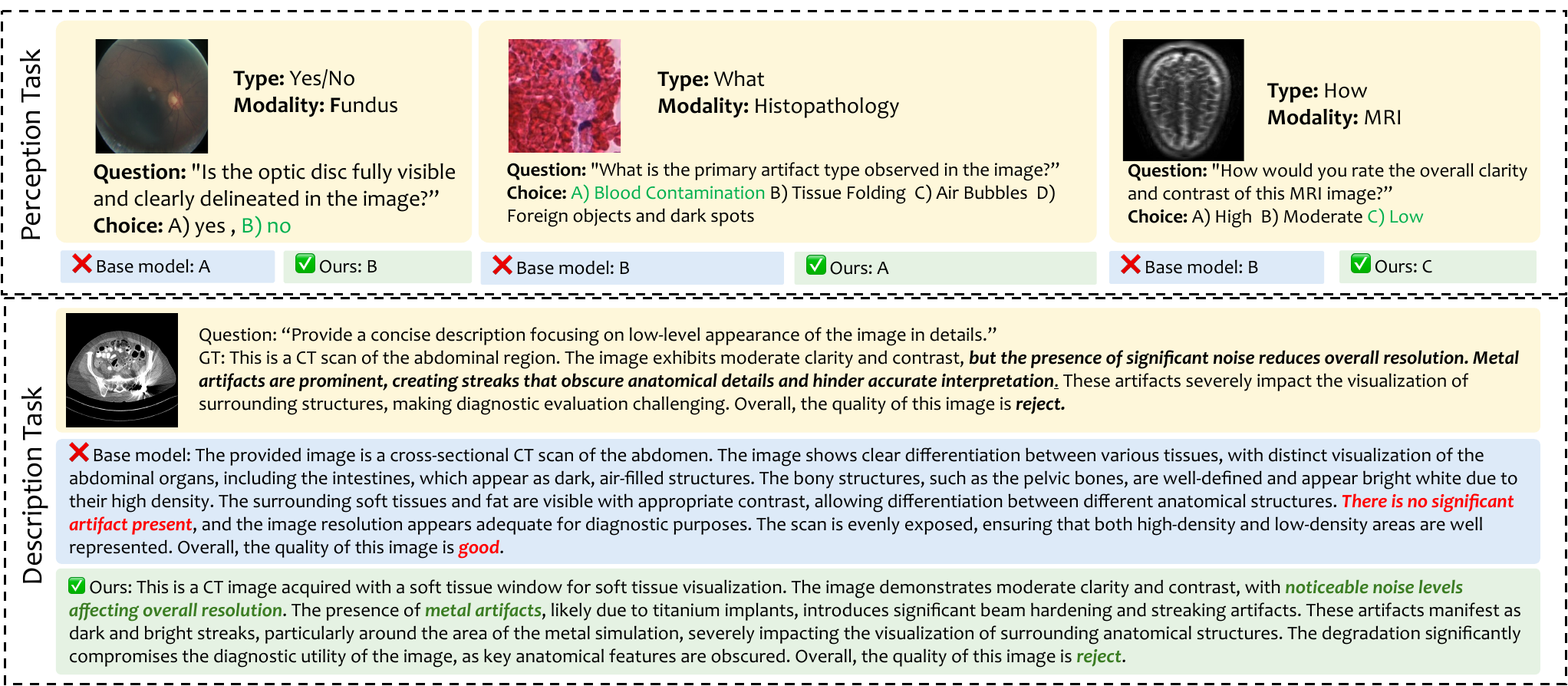}
    \caption{\textbf{Qualitative comparison} between InternVL3-8B (blue) and InternVL3-8B-10k$^\dagger$ (green).}
    \label{fig:case_study}
\end{figure*}

\textbf{Qualitative Case Studies.} Figure~\ref{fig:case_study} illustrates that, compared to base models which produce generic descriptions, MedQ-Engine generates anatomically specific assessments with clinical reasoning and actionable recommendations.

\textbf{Sample Efficiency.} We compare our failure-driven sampling against random sampling under identical annotation budgets. MedQ-Engine consistently outperforms random sampling across all data scales (Figure~\ref{fig:scaling}~(c)(f)). Notably, our method with only 10K samples surpasses random sampling at 40K samples and rapidly approaches the scaling upper bound, demonstrating more than 4$\times$ sample efficiency.

\textbf{Component Ablation.} Table~\ref{tab:ablation} isolates each component's contribution (InternVL3-8B, 10K samples). Human-in-the-loop verification contributes the largest gain, particularly critical for description quality. Capability-driven QA generation outperforms simple seed-rule QA, and adaptive sampling proves essential for prioritizing weak dimensions. Removing all engine components (random baseline) drops performance by 9.46\%, underscoring the cumulative value.

\textbf{Annotation Statistics.} Table~\ref{tab:annotation} summarizes annotation statistics across iterations. The high GPT-4o Accept rate (63\%) at cold-start reduces per-sample review time from 5.1 min (from scratch) to 0.5 min, a $10\times$ speedup. The progressive strategy further reduces average human review to only 18\% of samples in subsequent iterations, cutting overall expert cost by more than $5\times$ compared to full human review.

\begin{table}[t]
\begin{minipage}[t]{0.41\textwidth} 
    \centering
    \small
    \caption{Ablation study using InternVL3-8B with 10K samples.}
    \label{tab:ablation}
    \resizebox{\linewidth}{!}{
    \begin{tabular}{lcc}
    \toprule
    \textbf{Variant} & \textbf{Perc.$\uparrow$} & \textbf{Desc.$\uparrow$} \\
    \midrule
    Full MedQ-Engine & \textbf{78.16\%} & \textbf{5.65} \\
    \quad w/o human-in-the-loop & 73.25\% & 5.32 \\
    \quad w/o capability-driven QA & 75.43\% & 5.38 \\
    \quad w/o adaptive sampling & 74.87\% & 5.35 \\
    \quad w/o quality assurance & 76.51\% & 5.49 \\
    \quad Random sampling (10K) & 68.70\% & 5.28 \\
    \bottomrule
    \end{tabular}}
\end{minipage}
\hfill
\begin{minipage}[t]{0.575\textwidth}
    \centering
    \small
    \caption{Human-in-the-loop annotation statistics. $^\dagger$Selective review triggered by Sec.~\ref{sec:exploring}.}
    \label{tab:annotation}
    \resizebox{\linewidth}{!}{
    \begin{tabular}{lccc}
    \toprule
    \textbf{Metric} & \textbf{$t{=}0$} & \textbf{$t{>}0$} & \textbf{w/o pseudo-label} \\
    \midrule
    Annotation oracle       & GPT-4o        & Self $+$ GPT-4o & ---             \\
    Human review rate       & 100\%         & 18\%$^\dagger$  & 100\%           \\
    Accept rate      & 63\%          & ---             & ---             \\
    Edit rate        & 29\%          & ---             & ---             \\
    Reject rate             & 8\%           & ---             & ---             \\
    Avg.\ review time       & 0.5 min       & 0.5 min         & 5.1 min         \\
    \bottomrule
    \end{tabular}}
\end{minipage}
\end{table}

\subsection{Data Scaling Analysis}

\begin{figure*}[t]
    \centering
    \includegraphics[width=\textwidth]{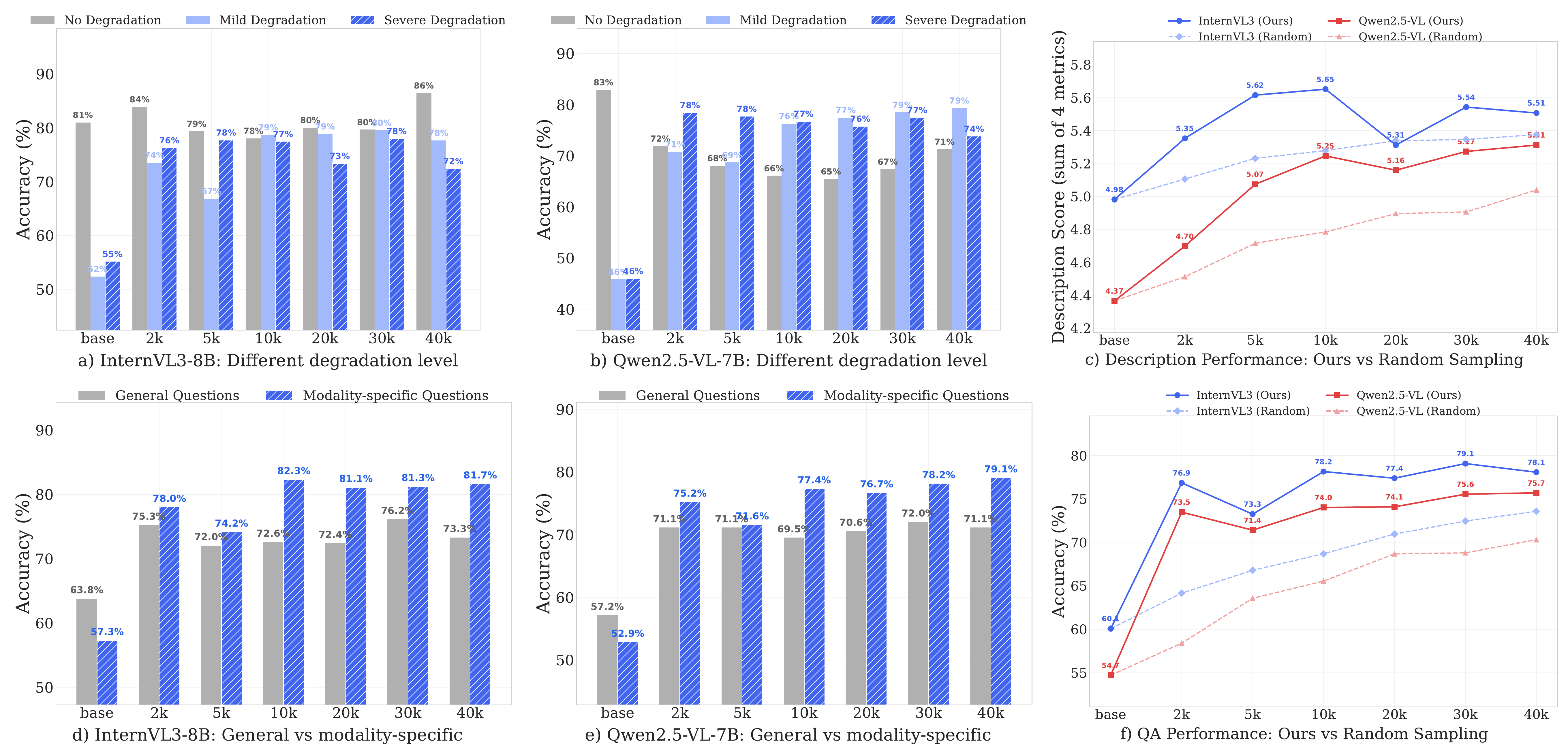}
    \caption{\textbf{Data scaling analysis across two model architectures.} \textit{Top row}: Performance on degradation severity levels and description scores as training data scales from 2K to 40K. \textit{Bottom row}: General vs.\ modality-specific question performance, and failure-driven vs.\ random sampling comparison.}
    \label{fig:scaling}
\end{figure*}

We investigate how model performance scales with training data size across different question types and degradation severity levels (Figure~\ref{fig:scaling}).

\textbf{Degradation Severity Analysis.} Base models perform well on artifact-free images but struggle with actual quality issues, where mild and severe degradation accuracy drops by over 30\% (Figure~\ref{fig:scaling}~(a)(b)). MedQ-Engine substantially narrows this gap, with the largest improvements on mild and severe cases, validating that failure-driven data collection effectively targets the most challenging scenarios.

\textbf{General vs.\ Modality-specific Questions.} After training, modality-specific questions consistently outperform general questions across both models (Figure~\ref{fig:scaling}~(d)(e)). This suggests that domain-specific training data enables models to develop specialized knowledge about modality-characteristic degradations, while general quality reasoning remains more challenging.

\section{Conclusion}
\label{sec:conclusion}

We present MedQ-Engine, a closed-loop data engine that systematically improves MLLM capabilities for medical image quality assessment. By clustering failure cases into prototypes, using them as retrieval anchors for targeted data expansion, and employing progressive human-in-the-loop annotation with entropy-guided selective review, MedQ-Engine reduces human involvement to 18\% of samples while achieving more than 4$\times$ sample efficiency over random sampling. With only 10K annotations, an 8B-parameter model surpasses GPT-4o by over 13\% and approaches human expert performance across five medical imaging modalities. Beyond Med-IQA, the evaluate-explore-evolve paradigm offers a general blueprint for data-efficient MLLM adaptation in specialized domains where expert annotations are scarce and model weaknesses are non-uniform.

\clearpage
\newpage

\bibliographystyle{splncs04}
\bibliography{bib}  

\end{document}